# Dynamic Toll Prediction Using Historical Data on Toll Roads: Case Study of the I-66 Inner Beltway

**Sara Zahedian**[*], **Amir Nohekhan, Kaveh Farokhi Sadabadi**
Center for Advanced Transportation Technology, University of Maryland, United States
* Corresponding author email: szahedi1@umd.edu

**Abstract**
Providing the users of a dynamic tolling system with predictions of tolling prices and the travel time difference between the toll road and the alternative routes enables them to make their travel decisions before starting their trip. This study aims to provide accurate predictions of tolling price through training and testing random forest, multilayer perceptron, and long short-term memory models and compare them with the current situation that the best prediction is extending the current toll to the next timesteps. The prediction time horizon includes five 6-minute time intervals ahead of the present time. The prediction performance of models over the testing set reveals that while all the models were significantly better than the base model, the random forest outperforms all models. For instance, while in the trained models, the mean absolute error range is from $1.5 to $2.5 for the next six minutes to the next 30 minutes, respectively, the same measure in the base model is in the range of $2.5 to $6. The prediction of travel time difference along the toll road and its alternative route with the shortest travel time revealed that the multilayer perceptron performs marginally better than the base model. However, due to a relatively stable travel time difference, the current travel time difference is an acceptable prediction for the next 30 minutes prediction horizon.

## 1. Introduction

This study aims to train and investigate various machine learning models to predict the toll price based on the historical data for facilities with dynamic tolling. Moreover, it examines the travel time difference between the toll facility and the alternative route with the shortest travel time for different prediction horizons to build a complete picture of the system's performance under the dynamic tolling scheme.

The implementation of dynamic tolling systems is an efficient manner for reducing traffic congestion (Zhang et al., 2008; Yin and Lou, 2009). In a dynamic tolling system, the price that users should pay to use the toll facility or enter a congested region is determined based on the travel demand at each particular time interval. In theory, the users alter their trip start time, mode choice, route choice, or other decisions based on the tolling amount (Button and Verhoef, 1998; Yang and Huang, 2005). Therefore, the effectiveness of a dynamic tolling system not only relies on an optimal tolling scheme (Zhu and Ukkusuri, 2014) but also on providing an accurate prediction of the tolling price and the time which can be saved using the toll facility to the users in advance (Mahmassani and Jayakrishnan, 1991; Noland, 1997; Knorr et al., 2014; Qi et al., 2020). This study contributes to the literature in predicting the toll price and travel time difference between the toll facility and the best alternative route for different timesteps without any knowledge of the operator's tolling algorithm. Therefore, the introduced models predict the amount of toll by extracting the interaction pattern between users and the operator. Such prediction is beneficial to users by reducing the uncertainty of their generalized travel costs. Moreover, such interactive communication between the users and the operator can minimize public opposition in implementing toll facilities (Odeck and Kjerkreit, 2010; Gu et al., 2018). Besides, it provides useful information to the system operator because unlike the actual toll set based on the existing algorithms, the predicted toll is estimated based on the historical data. Therefore, the operator can use it to gain insight into the users' interactions with the system in practice.

This study uses three well-known machine learning models, namely, Random Forest (RF), Multilayer Perceptron (MLP), and Long short-term memory (LSTM), to address the toll prediction problem. These models are trained using the historical data of the I-66 interstate inside the Capital Beltway, which is one of the most debating tolling facilities in the United States. This route is a High Occupancy Toll (HOT) facility during the rush hours in peak traffic direction. The dynamic toll fare is set every 6-minute to maintain a speed of 55 miles per hour along the corridor. This study uses the data of travel time, traffic flow, and the amount of toll over approximately 18 months and trains models that, given the data of the time t, predict the amount of toll for the next five 6-minute time intervals (i.e., t+6, t+12, …, t+30 minutes). The introduced models' performance is compared against the existing condition in which the best toll estimation available to the users is the current toll amount provided by the operator online[1]. Additionally, this study analyzes the travel time difference between the toll road and its alternative routes during the same prediction horizon to investigate the system performance and its predictability from the users' reaction perspective.

This paper is organized as follows. Section 2 reviews the related studies of the problem in transportation literature; then, we provide an overview of the study area in section 3. Section 4 explains the data sources used in this study and describe them for the study area. Section 5 provides more details about the methodology we used to address the toll prediction problem; whose numerical results are presented in Section 6. The last section summarizes the study framework and concludes the findings.

## 2. Literature Review

The ever-increasing demand for transportation in urban areas has long been a significant source for the transportation network's reduced service level. Transportation network pricing is a well-known strategy (Lindsey and Verhoef, 2001) and among the most successful travel demand management schemes that aim to use the existing infrastructure more efficiently by internalizing the external costs of congestion (Zhang et al., 2008; Yin and Lou, 2009). In addition to reducing congestion, another advantage of a pricing scheme is generating revenue for transportation authorities, which can be leveraged in improving the transportation supply and specifically the public transportation (Szeto and Lo, 2008; Zhu and Ukkusari, 2015). However, an optimal tolling strategy needs to conform to the users' time-dependent travel demand profile, which mandates a dynamic tolling scheme (Liu et al., 2009).

Road pricing strategies can be categorized based on various characteristics, such as the tolling scheme (de Palma and Lindsey, 2011). There are two primary road pricing schemes of cordon pricing and link-based pricing. In cordon pricing, vehicles entering a specified urban area should pay a cost, while in link-based pricing, vehicles traversing a road (or specific lanes of a road) should pay a toll. While cordon pricing is more prevalent in Europe and Asia, link-based pricing is more implemented in the US (Lou et al., 2011). Another approach to categorize road pricing is based on the time differentiation of the toll. Flat, time of day, responsive and anticipatory tolls are examples of such categorization (de Palma and Lindsey, 2011). Responsive and anticipatory dynamic tolling have entered the state of the practice more recently (Yin and Lou, 2009). The examples of such tolling are mostly limited to HOT facilities to maintain free-flow speed.

---

[1] https://vai66tolls.com/

The operator of the HOT lanes in a dynamic tolling strategy regulates the toll price to manage the single-occupancy vehicles (SOV) users of the road, thereby maintaining a desirable level-of-service (LOS) on the facility (Gardner et al., 2013). The SOV users decide whether to pay the toll and reduce travel time or use the regular facilities, typically experiencing a higher travel time. Therefore, the tolling price results from an interaction between the users and the system's operator, which in theory can improve the performance of both HOT and General-purpose (GP) lanes (Zhang et al., 2013). There are various techniques to implement dynamic tolling given the intelligent transportation systems (ITS) advancements in recent years. For a comprehensive up to date literature review of the studies in this area, readers may refer to Saharan et al. (2020).

While there are a handful of studies on the theoretical approaches to take for dynamic tolling at the design level, the number of researches on addressing the problems at an operational level is limited. The few studies in this area are mostly focused on descriptive analysis of the traffic condition in the locations with dynamic tolling on the HOT lanes such as I-394 MnPASS (Goodall and Smith, 2010) and I-15 San Diego (Borrell-Rovira and Supernak, 2011). However, to the best of the authors' knowledge, no study in the literature addresses the toll prediction problem. Therefore, this study contributes to the literature in predicting the toll price and travel time difference between the toll facility and the best alternative route for different prediction horizons.

There are several aspects of transportation operation and planning that have benefited from predicting the travel behavior and traffic flow characteristics (Sekula et al., 2018; Baee et al., 2019; Ahmed et al., 2020; Taghipour et al., 2020; Zahedian et al., 2020). Similarly, dynamic toll price prediction is beneficial to the users and operator since this variable impacts the travelers' behavior and, subsequently, the operator's response. From a user's viewpoint, this prediction is advantageous in estimating a more accurate cost for her trip, directly impacting the route choice, departure time, and mode choice. From the operator's perspective, providing toll prediction to the users can flatten the corridor's traffic volume profile, thereby reducing traffic congestion. Additionally, announcing toll predictions in advance can reduce the users' opposition since they have prior knowledge of the toll price and have made their decisions before starting their trip accordingly. From another perspective, the travel time difference between the tolled road and its alternative routes is the variable that indicates the responses of the users to the amount of toll. Therefore, predicting and analyzing the amount of toll and the travel time difference simultaneously is an essential step in addressing the dynamic tolling systems' operational challenges.

3. **Study area**

In the United States, the lanes of a highway are typically designated as regular or high-occupancy. The high-occupancy lanes are reserved for high-occupancy vehicles (HOV) and public transportation to promote carpooling and using transit. However, due to a generally low vehicle occupancy rate, the HOV lanes' capacity remains unused. The introduction of HOT lanes is to utilize the remaining capacity of the HOV lanes by allowing the SOVs to use the HOT lanes while paying a toll. One of the most debating HOT facilities in the United States is the I-66 Inner Beltway, which is the stretch of I-66 inside the Capital Beltway, I-495, located in the Washington DC metropolitan area.

The United States capital, Washington D.C., is situated between the states of Maryland and Virginia. Each day in the morning peak hours, commuters from the west side of Washington in Virginia use the Interstate-66 to arrive at their work location and do the opposite in the afternoon peak hours. Almost entirely located in Virginia, this freeway is the only east-west highway in the region, playing a significant role in its mobility. An overview of the study area in Figure 1 shows that I-66 crosses the Fairfax and Arlington counties and is adjacent to the independent cities of Fairfax and Falls Church. The entire corridor of I-66 is approximately 75 miles, which almost 11 miles lie inside the Capital Beltway. A comprehensive analysis of the I-66 corridor is presented at Nohekhan et al. (2021)

Due to the large volume of vehicles traversing the I-66 corridor on a daily basis, traffic congestion has been an ongoing issue on this corridor resulting in traffic jams, increased fuel consumption, and extended travel times. In the region, transportation authorities have implemented several congestion-relieving projects and proposed others to manage the travel demand and facilitate vehicles' movement. In one of the most recent projects, since 2017, the Virginia Department of Transportation designated the facility inside the Capital Beltway as HOT-2, allowing single-occupancy vehicles to pay the toll and use the corridor besides high-occupancy vehicles, which are exempt from this toll. Besides, all road users except motorcycles must have an EZ-Pass transponder, a small electronic device attached to the vehicle's windshield for paying tolls to traverse the segment. The tolling hours on the eastbound direction are from 5:30 to 9:30 AM and westbound from 3:00 to 7:00 PM. The toll price is set dynamically every 6 minutes based on several parameters such as the travel demand, traffic conditions, and the user's entrance to and exit location from the corridor. The tolling scheme aims at maintaining an average speed of 55 miles per hour, thus providing a reliable travel time to the users during the peak hours in the peak direction.

The project area and the location of on-ramps and off-ramps are illustrated in Figure 2. On each on- and off-ramp, installed toll gantries detect and determine the amount each user is charged. However, some of the ramps are grouped and have the same amount of toll.

The selection of a toll facility by a user depends upon various factors such as toll price, travel time that can be saved using the toll facility, travel time reliability, etc. Therefore, from the users' perspective, the travel time difference between the toll road and its competing routes is essential in decision-making. According to Google Maps' routing service, this study considers two alternative routes for the I-66 inside the beltway, one using Capital Beltway I-495 and George Washington Parkway (referred to as GW-PK in this paper), and the other one through US-50. These routes and their basic features, such as road length, number of lanes, and speed limit, are also illustrated in Figure 3.

## 4. Data Description

### 4.1. Data sources

This study uses several data sources to train and test models to predict dynamic toll prices and the travel time difference between the I-66 and its alternative routes. There three primary datasets used in this study are briefly described in what follows.

- **Toll data:** The complete data of I-66 Inner Beltway toll price based on time and entrance and exit locations is publicly available in XML format in the portal the SmarterRoads portal[1] maintained by VDOT. This dataset is continuously updated and incorporates the toll prices since June 2018.

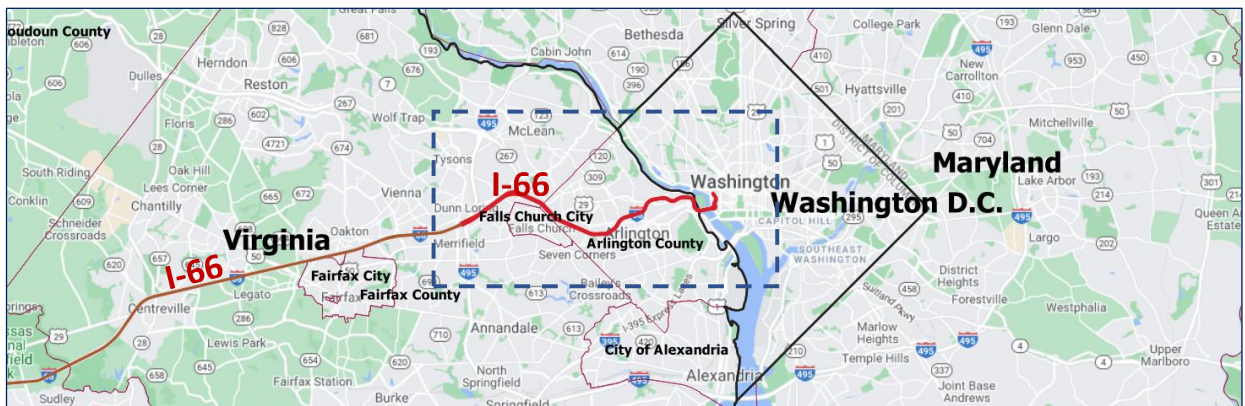

Figure 1 - I-66 location in the region (Nohekhan et al., 2021)

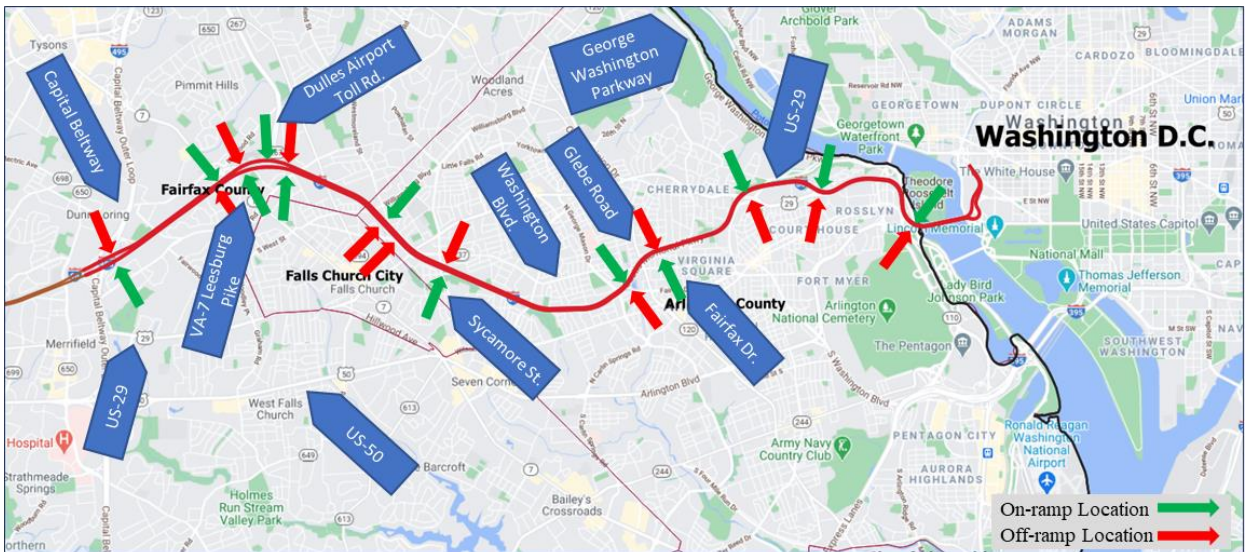

Figure 2 - I-66 inside the Capital Beltway and its ramps' locations (Nohekhan et al., 2021)

---

[1] https://smarterroads.org/

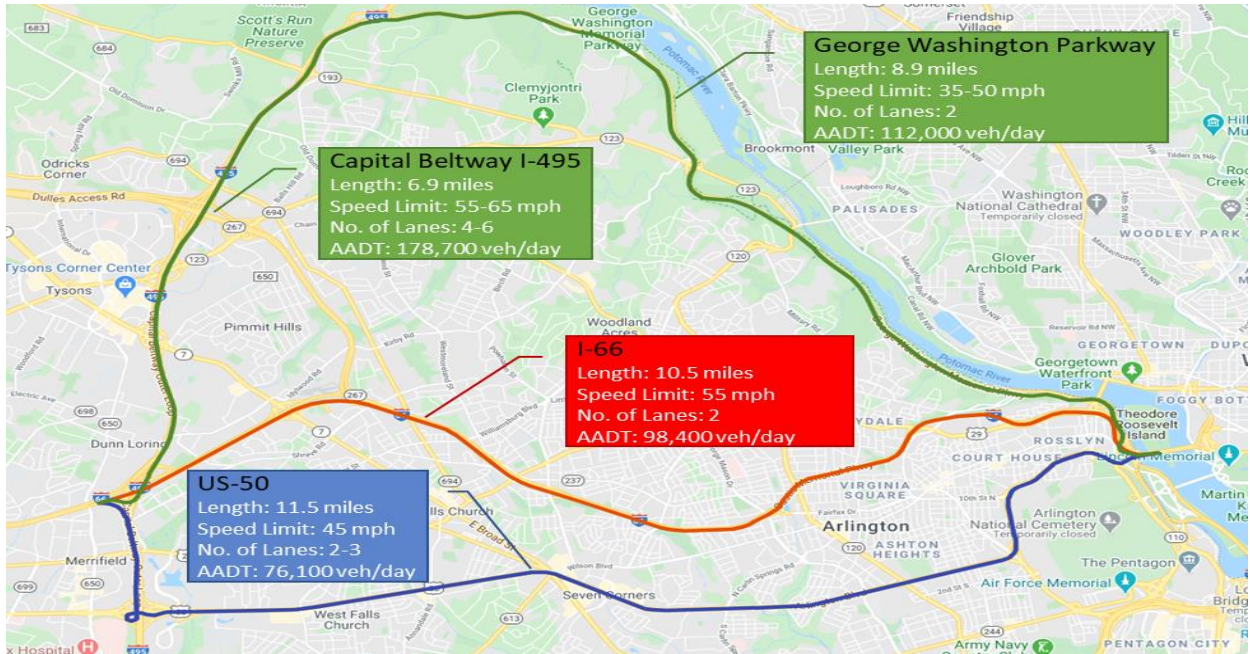
Figure 3 - I-66 and its alternative routes (GW-PK and US-50). (Nohekhan et al., 2021)

- **Probe Speed:** One of the features with high explainability of the traffic characteristics is the road segments' speed profile. The RITIS (Regional Integrated Transportation Information System) portal[1] holds the minute-by-minute speed data of TMC segments of the road network. In this portal, the speed for each segment is computed using the probe vehicle trajectory data. The number of TMC segments for each route at each direction is as follows:
    - I-66: 28 TMC segments
    - GW-PK: 22 TMC segments
    - US-50: 30 TMC segments
  
  The average speed along each route can be calculated using a weighted average of segments' speed and length. Besides, each route's travel time at each timestamp can be computed using the total length of the route divided by the average speed along the route at that timestamp.
- **Volume Counts:** In addition to toll and speed data, the lane-by-lane 15-minutes aggregated traffic volume data of one segment of the I-66 Inner Beltway is provided by the VDOT traffic monitoring program from the beginning of the year 2016 to the end of the year 2019. Although the available volume count data only belongs to the segment between VA-120 Glebe Road and US-29 Lee Highway, as shown in red in Figure 4, it can be used as a proxy for the I-66 Inner Beltway traffic volume.

After combining the mentioned datasets, the entire data covers the span of 18 months from the start of July 2018 to the end of December 2019 in 6-minutes time intervals.

### 4.2. Descriptive analysis

This section presents a descriptive analysis of the dataset with the aim of gaining more insight into the characteristics of the tolling scheme and the travel behavior of the users. As mentioned earlier, the tolling system of the I-66 inside the beltway is HOT lanes with dynamic tolling. As noted, the toll is collected from low occupancy vehicles traveling eastbound during the morning peak and traveling westbound during the afternoon peak. The toll price on this corridor changes with time (i.e., every 6-minute) and the locations where vehicles enter and exit the highway. Therefore, SOV users decide their route based on the toll price and the time savings by using the toll road. The distribution of toll price against travel time difference between the entire I-66 Inner Beltway corridor and its fastest alternative route during the tolling hours on the eastbound direction in 2018 and 2019 is illustrated in Figure 5(a). The heatmap of the same measures based on the number of observations is presented in Figure 5(b). Figures 6(a) and 6(b) illustrate the same measures for the westbound direction.

---
[1] https://ritis.org

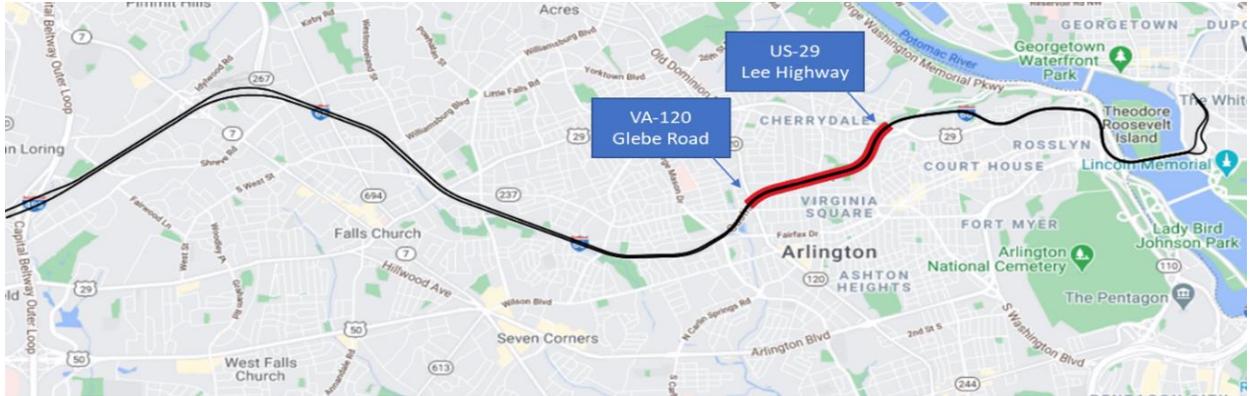

Figure 4 - Location of the segment with available traffic count data. (Google Maps)

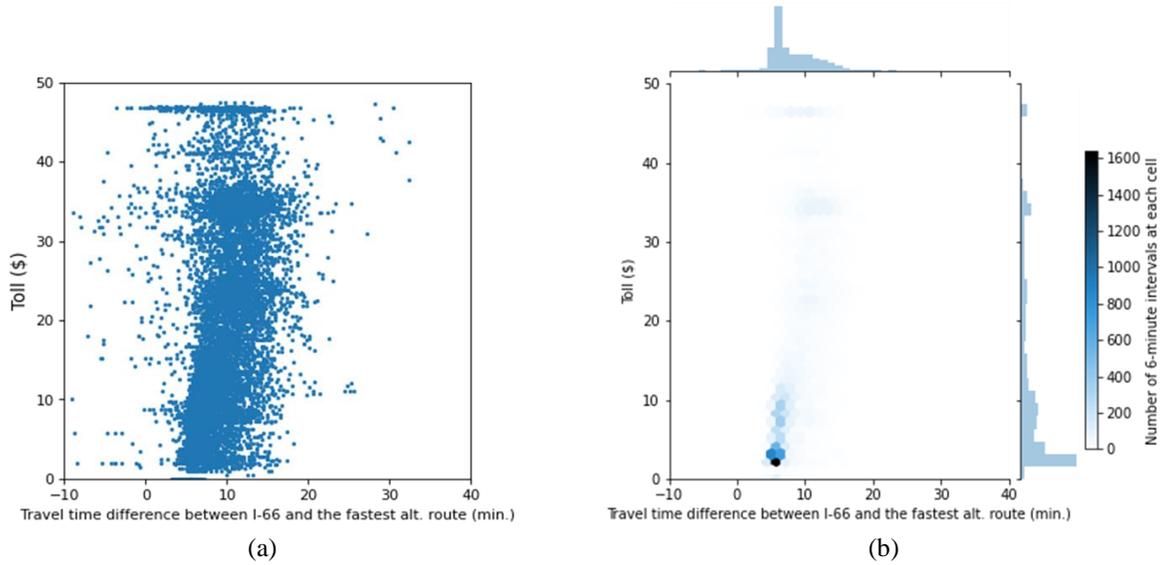

(a)          (b)

Figure 5 - Distribution of toll vs. travel time difference between I-66 and the fastest alternative route in the eastbound direction (a) data points (b) heatmap

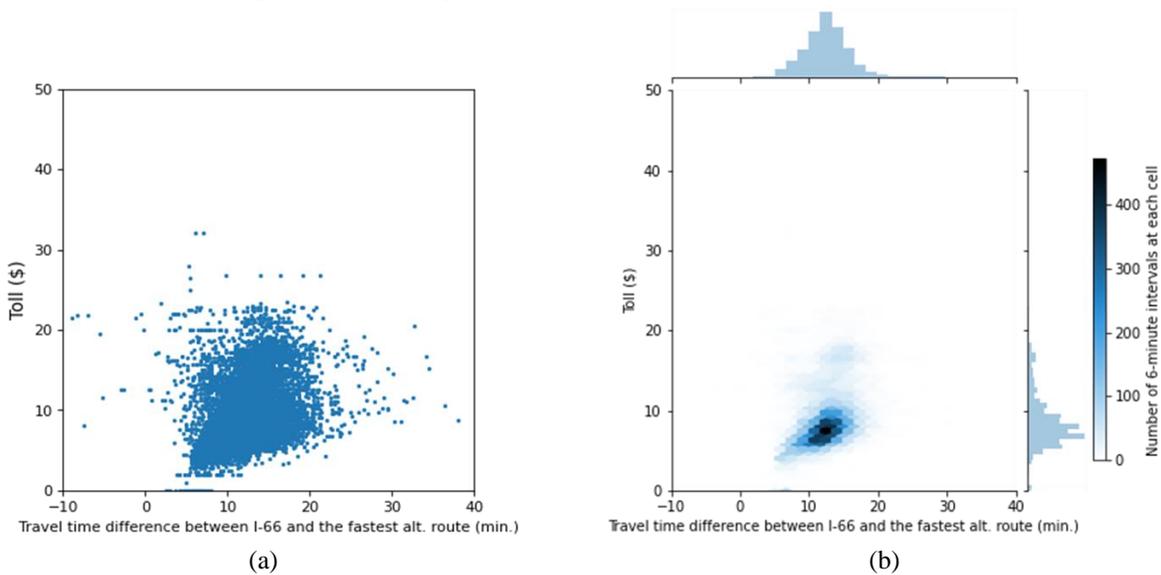

(a)          (b)

Figure 6 - Distribution of toll vs. travel time difference between I-66 and the fastest alternative route in the westbound direction (a) data points (b) heatmap

## 5. Methodology
### 5.1. Prediction models
This study employs several models to explore their performance in predicting toll on the I-66 HOT lanes inside the Capital Beltway in different timesteps ahead. In what follows, each of these models is described briefly. Interested readers may refer to Bonaccorso (2017) for more details on these algorithms.

#### 5.1.1. Random Forest
The random forest technique is an ensemble learning method constructed of many decision trees (Breiman, 2001; Rasouli and Timmermans, 2014; Song and Yin, 2015). A decision tree's basic concept is to break down a complex decision into several simple decisions (Safavian and Landgrebe, 1991; Myles et al., 2004). Unlike other approaches that jointly use a set of features in a single step, the decision tree is based on multistage (hierarchical) decision steps. In the random forest model, a subset of features is selected randomly to grow each decision tree. The random forest model result is obtained by a majority vote of the model's decision trees (Jahangiri and Rakha, 2015). The advantage of randomly selecting the subset of features is reducing the correlation between decision trees, which results in a reduced variance of errors (Breiman, 2001).

#### 5.1.2. Multilayer Perceptron (MLP)
The MLP model, as a feed-forward artificial neural network (ANN) class, is one of the most famous and widely used models for classification and regression problems. This model's general structure comprises of three types of layers: the input layer, hidden layers, and the output layer. These layers are mostly connected with nonlinear activation functions to a nonlinear model. The MLP model used in this study is constructed from four hidden layers connected with ELU activation functions. Batch normalization (Ioffe and Szegedy, 2015) and L2 regularization techniques (Cortes et al., 2012) are used in trained models to improve the performance. The loss function used in this study is the Mean Absolute Percentage Error (MAPE) optimized using the Adaptative Momentum (AdaM) optimizer (Kingma and Ba, 2014).

#### 5.1.3. Long Short-Term Memory (LSTM)
LSTM networks are a type of recurrent neural network (RNN) capable of learning order dependence in sequence prediction problems. Like other RNNs, LSTM is constructed from repeating ANNs; however, in LSTM, each chain's unit consists of a cell, an input gate, an output gate, and a forget gate. The three gates regulate the flow of information into the cell to avoid the long-term dependency problem of the simple RNN. For more details on the structure and formulation of the LSTM, interested readers may refer to the original work of Hochreiter and Schmidhuber (1997), where the authors introduced the algorithm for the first time.

In this study, the final finetuned LSTM model is constructed from one LSTM layer and three dense layers connected with the ELU activation function. As the MLP model, the MAPE loss function is used for this model and optimized with the AdaM algorithm.

### 5.2. Training and testing procedure
The prediction of toll and travel time difference helps the users decide their travel start time and route choice. As the toll price is dynamically set every six minutes, and travel times are time-dependent in their nature, this study trains and tests the prediction models for five 6-minute horizons of prediction ahead (i.e., 6, 12, 18, 24, 30) of each given time during the tolling hours. Thus, the output of the framework enables the users to have preliminary information on toll price and the time that can be saved using the toll road for different time intervals ahead, facilitating their decision making. The prediction performance metrics used in this study for comparing the models are Mean Absolute Error (MAE), MAPE, and R-squared. Each of these metrics illustrates a different aspect of prediction performance. The MAE represents the errors in terms of monetary values or travels time depending on the output. The MAPE measures the percentage of error in prediction, and R-square illustrates the model's overall performance relative to the average values.

Before training the introduced models, the dataset is divided into three sections: training, validation, and test sets. Each model is trained on the training set, and the hyperparameters are determined based on evaluating the model on the validation set. Lastly, the models are compared in their prediction performance on the test set. The validation set is the data of the last three weeks of the dataset in 2019, and the testing set is the data of two randomly selected days from each month of the entire data yielding 36 days. The rest of the dataset is used for training the models.

The base model for comparing the model performances is the framework in which the current value of toll or travel time difference is reported for the timesteps ahead without making any changes. This framework represents the current situation where the best prediction that users can make is extending the reported toll price to later timesteps.

## 6. Results and Discussion
This section presents the results of training and testing the models. The first target value is toll price at five timesteps ahead, each with their own model trained and tested. The performance measures introduced earlier are computed on the training and testing sets. Figure 7 compares the MAE, MAPE, and $R^2$ for all four models in different prediction horizons. According to this figure, intuitively, the average absolute error for all models increase as the prediction is farther away in time. However, in all timesteps, the base model's error is much higher than other models, and it reaches a difference of six dollars when predicting for 30 minutes ahead. The same measure for all other models is less than three dollars in 30 minutes ahead. The random forest model is outperforming the MLP and LSTM models in all timesteps, although slightly. As far as MAPE value is concerned, the RF, MLP, and LSTM models have a MAPE of less than 20 percent for all timesteps ahead, while this measure for the base model is much higher, starting at more than 30 percent and exceeding 70 percent in 30 minutes. Contrary to the previous measure, here, the MLP model presented the best performance with the least MAPE for all timesteps. This performance is not unexpected from the MLP model since its objective function is minimizing MAPE. Figure 7 also shows the $R^2$ values of different models in different prediction horizons. According to this figure, except for the base model, all other models have an $R^2$ of more than 0.9 in all timesteps. However, going farther, R-square reduces over the testing set for all models as expected. In terms of model goodness-of-fit, the random forest model is superior to other models, although slightly except for predicting toll in 18 minutes ahead of the current time when the MLP model performs better.

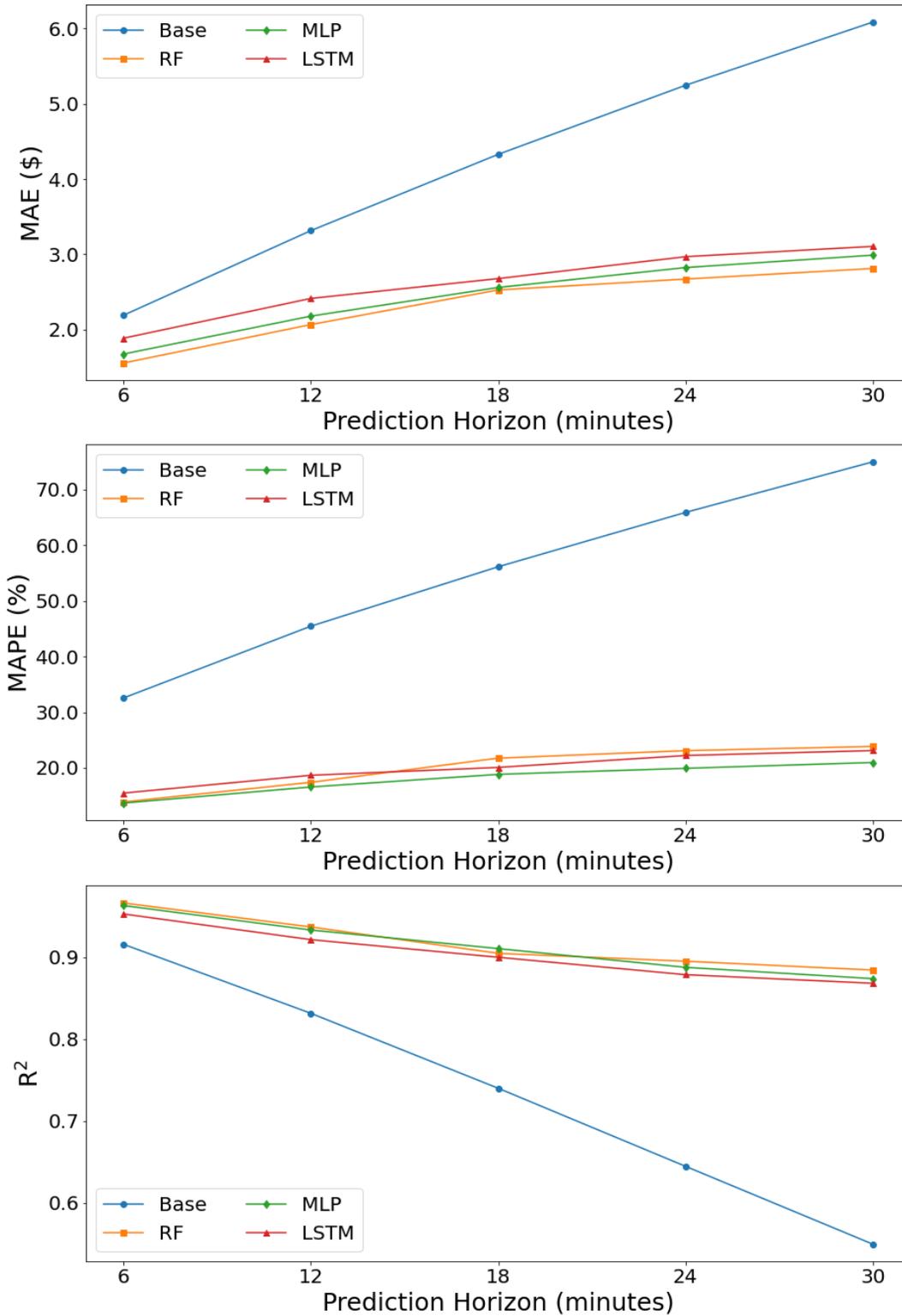

Figure 7 – Comparison of models' MAPE, MAE and $R^2$ for toll price prediction in different prediction horizons.

While the above-illustrated metrics are all aggregated measures for the models' overall performance, they are not representative of the prediction error variations. Therefore, the variations of prediction errors are illustrated in the box and whisker plots in Figure 8. Based on this figure, the median difference between actual and predicted toll price, as a measure of central tendency, is lowest in the random forest (almost equal to zero), followed by MLP, LSTM, and the base model. The random forest model is also outperforming other models in terms of variations of toll prediction errors. Besides, it is clear that the farther timesteps result in higher variations in errors for all models. Moreover, the random forest model is capable of predicting the toll price with almost 2 dollars difference in 50 percent of times and approximately 4 dollars in all times. In comparison, this value can be as much as 12.5 dollars when users only have access to the online toll price.

As stated throughout the paper, one other perspective to evaluate a dynamic tolling system's performance is to investigate the travel time difference between the toll road and its alternative routes. This section provides the results of travel time difference prediction for I-66 and its alternative routes. Here we only present the prediction results for the best model, which is the MLP in the case of travel time difference prediction, to focus on this variable's characteristics rather than comparing various prediction models. Figure 9 compares the MLP model and the base model (i.e., the travel time difference's value at the prediction timestep 0) in terms of MAE, MAPE, and $R^2$. In addition, the Box and Whisker plots of travel time difference prediction errors for the Base and MLP models in different prediction horizons are illustrated in Figure 10. According to these figures, the MLP provides considerably lower MAPE and MAE values, especially as we get farther in the prediction horizon. However, looking at the $R^2$ values in Figure 9 and the variation of prediction error in Figure 10, it can be induced that the prediction model is not providing significant improvement compared to the base model. This lack of substantial improvement is because the travel time difference between the I-66 and its alternative routes does not significantly vary over time (refer to Figures 5 and 6). Therefore, the average value of travel time difference itself can be an acceptable estimation leading to low $R^2$ values for the MLP model.

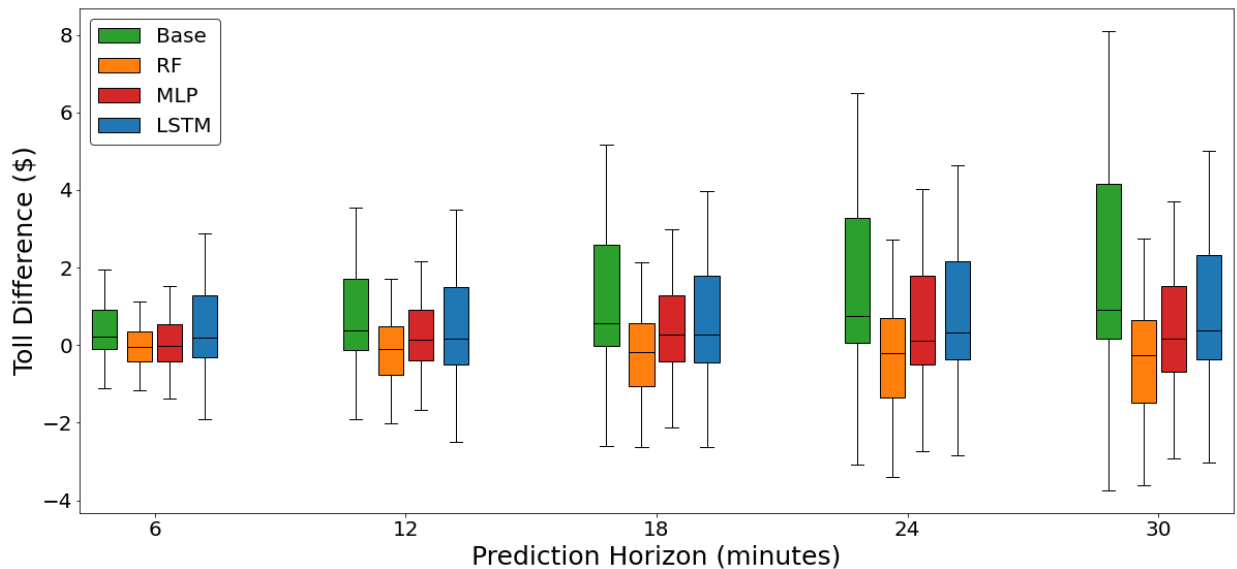

Figure 8. Comparison of Box and Whisker plots of toll price prediction errors in different prediction horizons.

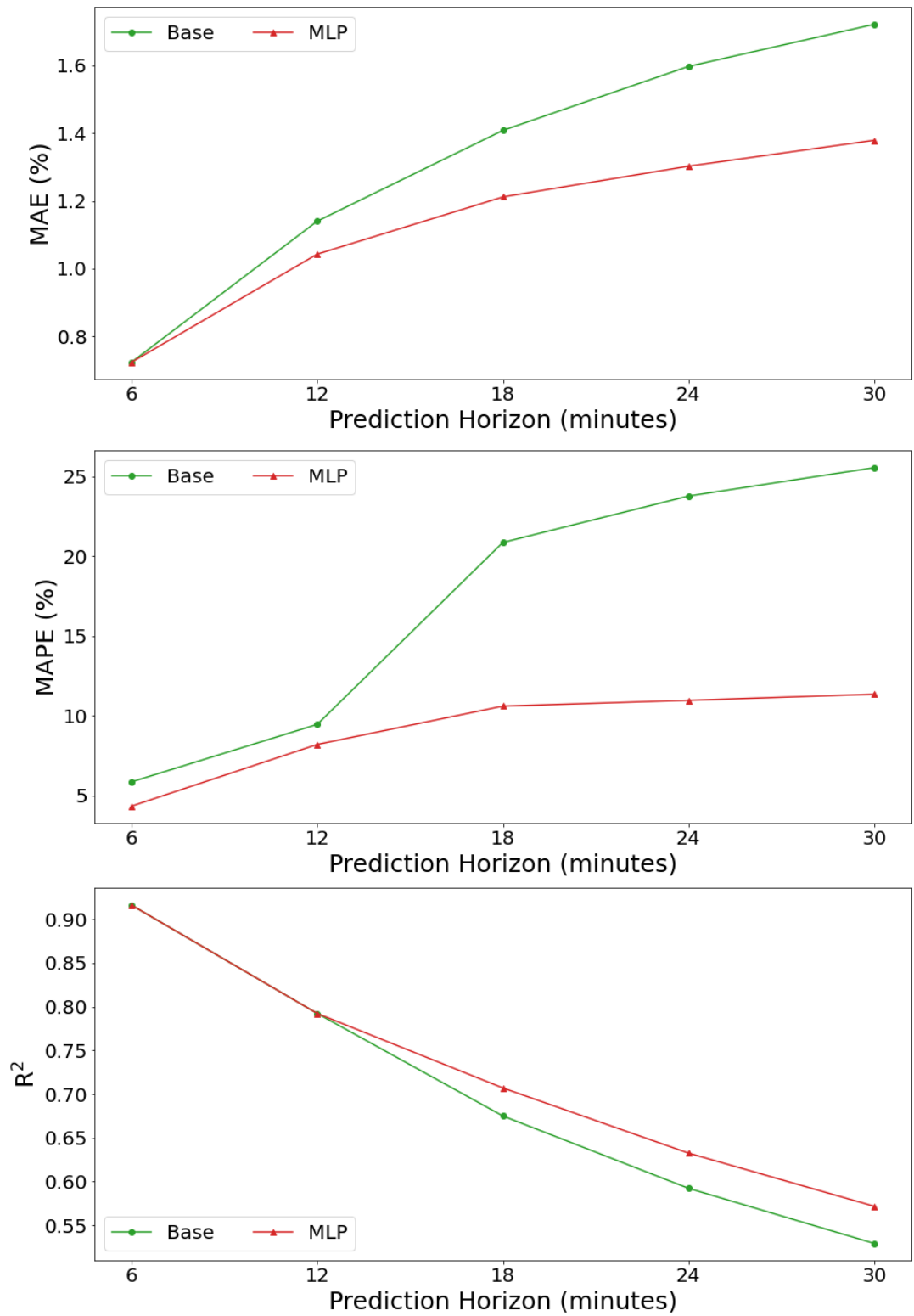

Figure 9 – Comparison of the Base and MLP models' MAPE, MAE and $R^2$ for travel time difference prediction in different prediction horizons.

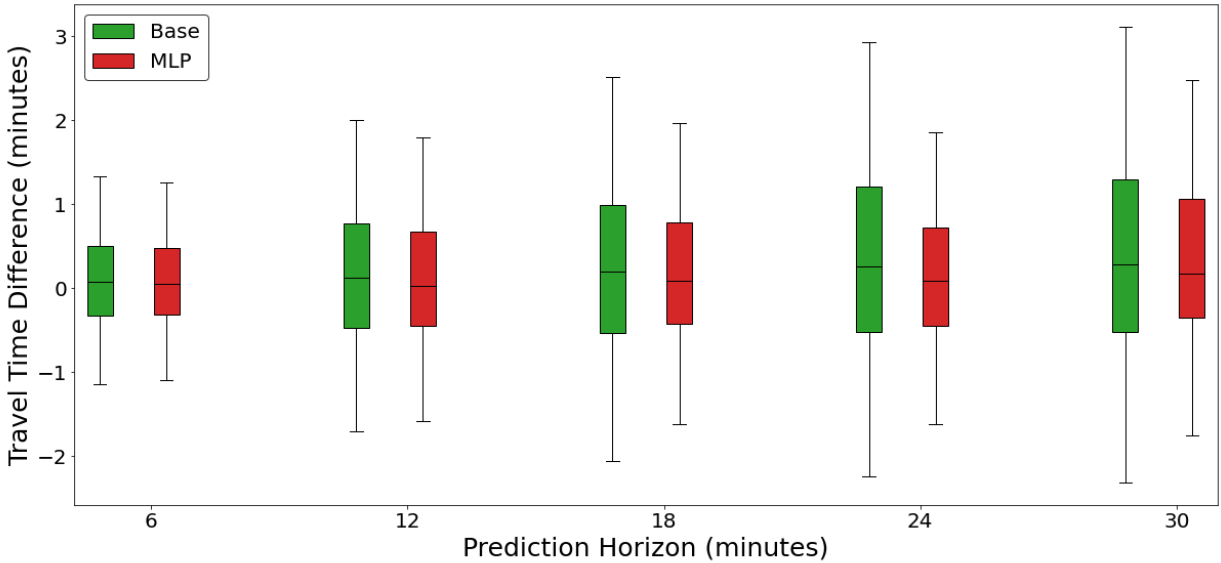

Figure 10. Box and Whisker plots of travel time difference prediction errors for the Base and MLP models in different prediction horizons.

## 7. Summary and Conclusion

This study investigates the predictability of toll in a dynamic tolling scheme using historical data of the probe travel time, sparse loop detectors, and the amount of toll. Three well-known machine learning algorithms of RF, MLP, and LSTM are trained and tested to predict the amount of toll in the I-66 for five timesteps in a prediction horizon of 30 minutes. The models' inputs are travel conditions at the current timestep, including travel time, traffic volume, and toll. All three models are compared together and to a base model, which is defined based on the current situation that tolling information is only available for the current timestep.

The results indicate that all three models have acceptable performance in predicting the toll price. However, overall, RF outperforms the other two models slightly. Intuitively, the performance of models degrades as we go further in time. Quantitatively, the RF model average toll prediction error (i.e., MAE) ranges from $1.5 for the next 6 minutes and $2.5 for the next 30 minutes, while these numbers for the base model are approximately from $2.5 to $6 respectively. Dividing these values to the actual amount of toll yields the MAPE value, which is less than 20% for prediction models over the entire prediction horizon. In comparison, this value rises to approximately 75% for the base model. Another metrics that are used to compare the performance of the models is $R^2$. The same as the other two metrics, $R^2$ values confirm the capabilities of the prediction models. This value is more than 90% for the RF model over the entire prediction horizon.

In addition to toll prediction, this study investigates the predictability of travel time difference between the study toll road (i.e., I-66) and its alternative routes. The results reveal that although models such as MLP can predict this value with acceptable accuracy, their performance is not significantly better than the base model. This stems from the fact that although the travel time on each route varies over time, the travel time difference between the toll road and alternative routes in this study area is relatively stable during the tolling hours. Thus, the current time travel time difference itself is an acceptable approximation for all prediction horizons over the next 30 minutes.

This study's findings can be used in practice to provide users with a relatively accurate estimation of the toll. This information helps the driver plan their trip time and travel mode having a more reliable estimation of the trip cost. The introduced framework is directly applicable to any other dynamic toll road whose historical data is available. While the current study uses limited data sources to solve the prediction model, more comprehensive datasets may improve toll prediction accuracy. Adding information such as the weather condition, accidents in the area, and more granular traffic volume data will be investigated in future works.

### Acknowledgements

The authors would like to thank the organizers of the Student Data Challenge on Urban Travel Time, Speed, and Reliability, held in 2019, for providing the data used in this paper through the VDOT Smarter Roads Platform.

**References**

Ahmed, Tanjeeb, Michael Hyland, Navjyoth JS Sarma, Suman Mitra, and Arash Ghaffar. "Quantifying the employment accessibility benefits of shared automated vehicle mobility services: Consumer welfare approach using logsums." Transportation Research Part A: Policy and Practice 141 (2020): 221-247.

Baee, Sonia, Erfan Pakdamanian, Vicente Ordonez Roman, Inki Kim, Lu Feng, and Laura Barnes. "EyeCar: Modeling the Visual Attention Allocation of Drivers in Semi-Autonomous Vehicles." arXiv preprint arXiv:1912.07773 (2019).

Bonaccorso, Giuseppe. Machine learning algorithms. Packt Publishing Ltd, 2017.

Borrell-Rovira, Anna, and Janusz Supernak. "Peak Period Use of I-15 Corridor in San Diego, California: Long-Term Impact of FasTrak Program." Transportation research record 2221, no. 1 (2011): 64-73.

Breiman, Leo. "Random forests." Machine learning 45.1 (2001): 5-32.

Button, Kenneth, and Erik Verhoef. Road pricing, traffic congestion and the environment. Edward Elgar Publishing, 1998.

Cortes, Corinna, Mehryar Mohri, and Afshin Rostamizadeh. "L2 regularization for learning kernels." arXiv preprint arXiv:1205.2653 (2012).

de Palma, André, and Robin Lindsey. "Traffic congestion pricing methodologies and technologies." Transportation Research Part C: Emerging Technologies 19, no. 6 (2011): 1377-1399.

Gers, Felix A., Jürgen Schmidhuber, and Fred Cummins. "Learning to forget: Continual prediction with LSTM." (1999): 850-855.

Goodall, Noah, and Brian L. Smith. "What drives decisions of single-occupant travelers in high-occupancy vehicle lanes? Investigation using archived traffic and tolling data from MnPASS express lanes." Transportation research record 2178, no. 1 (2010): 156-161.

Gu, Ziyuan, Zhiyuan Liu, Qixiu Cheng, and Meead Saberi. "Congestion pricing practices and public acceptance: A review of evidence." *Case Studies on Transport Policy* 6, no. 1 (2018): 94-101.

Hochreiter, Sepp, and Jürgen Schmidhuber. "Long short-term memory." Neural computation 9.8 (1997): 1735-1780.

Ioffe, S. and Szegedy, C., 2015. Batch normalization: Accelerating deep network training by reducing internal covariate shift. arXiv preprint arXiv:1502.03167.

Jahangiri, Arash, and Hesham A. Rakha. "Applying machine learning techniques to transportation mode recognition using mobile phone sensor data." IEEE transactions on intelligent transportation systems 16.5 (2015): 2406-2417.

Kingma, D.P. and Ba, J., 2014. Adam: A method for stochastic optimization. arXiv preprint arXiv:1412.6980.

Knorr, Florian, Thorsten Chmura, and Michael Schreckenberg. "Route choice in the presence of a toll road: The role of pre-trip information and learning." *Transportation Research Part F: Traffic Psychology and Behaviour* 27 (2014): 44-55.

Lindsey, R., and E. Verhoef. "Traffic Congestion and Congestion pricing. in: Handbook of Transport Systems and Traffic Control." Publication of: Elsevier Scientific Publishing Company (2001).

Liu, Tian-liang, Hai-jun Huang, and Li-jun Tian. "Microscopic simulation of multi-lane traffic under dynamic tolling and information feedback." Journal of Central South University of Technology 16.5 (2009): 865.

Lou, Yingyan, Yafeng Yin, and Jorge A. Laval. "Optimal dynamic pricing strategies for high-occupancy/toll lanes." Transportation Research Part C: Emerging Technologies 19, no. 1 (2011): 64-74.

Mahmassani, Hani S., and R. Jayakrishnan. "System performance and user response under real-time information in a congested traffic corridor." *Transportation Research Part A: General* 25, no. 5 (1991): 293-307.

Myles, Anthony J., et al. "An introduction to decision tree modeling." Journal of Chemometrics: A Journal of the Chemometrics Society 18.6 (2004): 275-285.

Nohekhan, Amir, Sara Zahedian, and Kaveh Farokhi Sadabadi. "Investigating the impacts of I-66 Inner Beltway dynamic tolling system." Transportation Engineering 4 (2021): 100059.

Noland, Robert B. "Commuter responses to travel time uncertainty under congested conditions: expected costs and the provision of information." *Journal of urban economics* 41, no. 3 (1997): 377-406.

Odeck, James, and Anne Kjerkreit. "Evidence on users' attitudes towards road user charges—A cross-sectional survey of six Norwegian toll schemes." *Transport Policy* 17, no. 6 (2010): 349-358.

Qi, Hang, Shoufeng Ma, Ning Jia, and Guangchao Wang. "Individual response modes to pre-trip information in congestible networks: laboratory experiment." *Transportmetrica A: Transport Science* 15, no. 2 (2019): 376-395.

Rasouli, Soora, and Harry JP Timmermans. "Using ensembles of decision trees to predict transport mode choice decisions: Effects on predictive success and uncertainty estimates." European Journal of Transport and Infrastructure Research 14.4 (2014).

Safavian, S. Rasoul, and David Landgrebe. "A survey of decision tree classifier methodology." IEEE transactions on systems, man, and cybernetics 21.3 (1991): 660-674.

Saharan, Sandeep, Seema Bawa, and Neeraj Kumar. "Dynamic pricing techniques for Intelligent Transportation System in smart cities: A systematic review." Computer Communications 150 (2020): 603-625.

Sekuła, Przemysław, Nikola Marković, Zachary Vander Laan, and Kaveh Farokhi Sadabadi. "Estimating historical hourly traffic volumes via machine learning and vehicle probe data: A Maryland case study." Transportation Research Part C: Emerging Technologies 97 (2018): 147-158.

Shi, Xingjian, et al. "Convolutional LSTM network: A machine learning approach for precipitation nowcasting." Advances in neural information processing systems 28 (2015): 802-810.

Song, Yan-Yan, and L. U. Ying. "Decision tree methods: applications for classification and prediction." Shanghai archives of psychiatry 27.2 (2015): 130.

Szeto, Wai Yuen, and Hong K. Lo. "Time-dependent transport network improvement and tolling strategies." Transportation Research Part A: Policy and Practice 42.2 (2008): 376-391.

Taghipour, Homa, Amir Bahador Parsa, and Abolfazl Kouros Mohammadian. "A dynamic approach to predict travel time in real time using data driven techniques and comprehensive data sources." Transportation Engineering 2 (2020): 100025.

Yang, Hai, and Hai-Jun Huang. *Mathematical and economic theory of road pricing*. 2005.

Yin, Yafeng, and Yingyan Lou. "Dynamic tolling strategies for managed lanes." Journal of Transportation Engineering 135.2 (2009): 45-52.

Zahedian, Sara, Przemysław Sekuła, Amir Nohekhan, and Zachary Vander Laan. "Estimating Hourly Traffic Volumes using Artificial Neural Network with Additional Inputs from Automatic Traffic Recorders." Transportation Research Record 2674, no. 3 (2020): 272-282.

Zhang, Guohui, Xiaolei Ma, and Yinhai Wang. "Self-adaptive tolling strategy for enhanced high-occupancy toll lane operations." IEEE Transactions on Intelligent Transportation Systems 15, no. 1 (2013): 306-317.

Zhang, Guohui, Yinhai Wang, Heng Wei, and Ping Yi. "A feedback-based dynamic tolling algorithm for high-occupancy toll lane operations." *Transportation Research Record* 2065, no. 1 (2008): 54-63.

Zhu, Feng, and Satish V. Ukkusuri. "A reinforcement learning approach for distance-based dynamic tolling in the stochastic network environment." Journal of Advanced Transportation 49.2 (2015): 247-266.